\documentclass{svmult}
\usepackage{helvet}
\usepackage{courier}
\usepackage[bottom]{footmisc}
\usepackage[pdfstartview={FitH},bookmarks=false]{hyperref}
\usepackage{url}
\usepackage{breakurl}
\usepackage{graphicx,color}
\usepackage{paralist}
\usepackage{amsmath,amsfonts,amssymb}
\usepackage{subfigure}
\usepackage{algorithm2e}
\usepackage{multirow}
\usepackage{multicol}
\usepackage{fancyhdr}

\newcommand{\ul}{\underline}

\newcommand{\abs}[1]{\left\lvert #1 \right\rvert}
\newcommand{\norm}[1]{\left\lVert #1 \right\rVert}
\newcommand{\card}[1]{\abs{#1}}
\newcommand{\vect}[1]{\ensuremath{\mathbf{#1}}}
\newcommand{\set}[1]{\ensuremath{\mathcal{#1}}}
\DeclareMathOperator*{\argmax}{arg\,max}
\DeclareMathOperator*{\argmin}{arg\,min}

\begin{document}

\title*{Linear, Deterministic, and Order-Invariant Initialization Methods for the K-Means Clustering Algorithm}
\titlerunning{Deterministic Initialization of the K-Means Algorithm}

\author{M.\ Emre Celebi \and Hassan A.\ Kingravi}
\institute{%
        M. Emre Celebi \at
        Department of Computer Science\\Louisiana State University, Shreveport, LA, USA\\
        \email{ecelebi@lsus.edu}\\
        \and
        Hassan A.\ Kingravi \at
        School of Electrical and Computer Engineering\\Georgia Institute of Technology, Atlanta, GA, USA\\
        \email{kingravi@gatech.edu}\\
       }

\maketitle

\abstract
{
Over the past five decades, k-means has become the clustering algorithm of choice in many application domains primarily due to its simplicity, time/space efficiency, and invariance to the ordering of the data points. Unfortunately, the algorithm's sensitivity to the initial selection of the cluster centers remains to be its most serious drawback. Numerous initialization methods have been proposed to address this drawback. Many of these methods, however, have time complexity superlinear in the number of data points, which makes them impractical for large data sets. On the other hand, linear methods are often random and/or sensitive to the ordering of the data points. These methods are generally unreliable in that the quality of their results is unpredictable. Therefore, it is common practice to perform multiple runs of such methods and take the output of the run that produces the best results. Such a practice, however, greatly increases the computational requirements of the otherwise highly efficient k-means algorithm. In this chapter, we investigate the empirical performance of six linear, deterministic (non-random), and order-invariant k-means initialization methods on a large and diverse collection of data sets from the UCI Machine Learning Repository. The results demonstrate that two relatively unknown hierarchical initialization methods due to Su and Dy outperform the remaining four methods with respect to two objective effectiveness criteria. In addition, a recent method due to Eri\c{s}o\u{g}lu \emph{et al.} performs surprisingly poorly.
}

\section{Introduction}
\label{sec_intro}
Clustering, the unsupervised classification of patterns into groups, is one of the most important tasks in exploratory  data analysis \cite{Jain99}. Primary goals of clustering include gaining insight into, classifying, and compressing data. Clustering has a long and rich history in a variety of scientific disciplines including anthropology, biology, medicine, psychology, statistics, mathematics, engineering, and computer science. As a result, numerous clustering algorithms have been proposed since the early 1950s \cite{Jain10}.

Clustering algorithms can be broadly classified into two groups: hierarchical and partitional \cite{Jain99}. Hierarchical algorithms recursively find nested clusters either in a top-down (divisive) or bottom-up (agglomerative) fashion. In contrast, partitional algorithms find all the clusters simultaneously as a partition of the data and do not impose a hierarchical structure. Most hierarchical algorithms have time complexity quadratic or higher in the number of data points \cite{Xu05} and therefore are not suitable for large data sets, whereas partitional algorithms often have lower complexity.

Given a data set $\set{X} = \{ \vect{x}_1, \vect{x}_2, \dotsc, \vect{x}_N \} \subset \mathbb{R}^D$, i.e., $N$ points (vectors) each with $D$ attributes (components), hard partitional algorithms divide $\set{X}$ into $K$ exhaustive and mutually exclusive clusters $\set{C} = \{ \set{C}_1, \set{C}_2, \dotsc, \set{C}_K \},$ $\quad \bigcup\nolimits_{i = 1}^K {\set{C}_i = \set{X}},$ $\quad \set{C}_i \cap \set{C}_j = \varnothing$ for $1 \leq i \neq j \leq K$. These algorithms usually generate clusters by optimizing a criterion function~\cite{Hall12}. The most intuitive and frequently used criterion function is the Sum of Squared Error (SSE) given by
\begin{equation}
\label{eq_sse}
\text{SSE} = \sum\limits_{i = 1}^K {\sum\limits_{\vect{x}_j \in \set{C}_i} {\norm{\vect{x}_j - \vect{c}_i}_2^2 } } ,
\end{equation}
where
\begin{equation}
\vect{c}_i =
\dfrac{1}{\card{\set{C}_i}}
\sum_{\vect{x}_j \in \set{C}_i} {\vect{x}_j}
\end{equation}
and
\begin{equation}
\norm{\vect{x}_j}_2 = \left( \sum_{d = 1}^D x_{jd}^2 \right)^{1/2}
\end{equation}
denote the centroid of cluster $\set{C}_i$ (with cardinality $\card{\set{C}_i}$) and the Euclidean ($\ell_2$) norm of vector $\vect{x}_j = (x_{j1}, x_{j2}, \dotsc, x_{jD})$, respectively.

The number of ways in which a set of $N$ objects can be partitioned into $K$ non-empty groups is given by Stirling numbers of the second kind
\begin{equation}
\mathcal{S}(N,K) =
\frac{1}{K!}
\sum_{i = 0}^K {(-1)^{K - i} \binom{K}{i} i^N} ,
\end{equation}
which can be approximated by $K^N/K!$ It can be seen that a complete enumeration of all possible clusterings to determine the global minimum of \eqref{eq_sse} is clearly computationally prohibitive except for very small data sets. In fact, this non-convex optimization problem is proven to be NP-hard even for $K = 2$ \cite{Dasgupta08,Aloise09} or $D = 2$ \cite{Vattani11,Mahajan12}. Consequently, various heuristics have been developed to provide approximate solutions to this problem \cite{Tarsitano03}. Most of the early approaches \cite{Forgy65,Jancey66,MacQueen67,Sparks73,Spath77,Banfield77,Hartigan79,Lloyd82} were simple procedures based on the alternating minimization algorithm \cite{Csiszar84}. In contrast, recent approaches are predominantly based on various metaheuristics \cite{RaywardSmith05,Das09} that are capable of avoiding bad local minima at the expense of significantly increased computational requirements. These include heuristics based on simulated annealing \cite{Klein89}, evolution strategies \cite{Babu94b}, tabu search \cite{AlSultan95}, genetic algorithms \cite{Murty96}, variable neighborhood search \cite{Hansen01}, memetic algorithms \cite{Pacheco03}, scatter search \cite{Pacheco05}, ant colony optimization \cite{Handl05}, differential evolution \cite{Paterlini06}, and particle swarm optimization \cite{Paterlini06}. Among all these heuristics, Lloyd's algorithm \cite{Lloyd82}, often referred to as the (batch) k-means algorithm, is the simplest and most commonly used one. This algorithm starts with $K$ arbitrary centers, typically chosen uniformly at random from the data points. Each point is assigned to the nearest center and then each center is recalculated as the mean of all points assigned to it. These two steps are repeated until a predefined termination criterion is met. K-means can be expressed in algorithmic notation as follows:

\begin{enumerate}
  \item Choose the initial set of centers $\vect{c}_1, \vect{c}_2, \dotsc, \vect{c}_K$ arbitrarily.
  \item Assign point $\vect{x}_j$ ($j \in \{ 1, 2, \dotsc, N \}$) to the nearest center with respect to $\ell_2$ distance, that is
        \begin{equation*}
        \vect{x}_j \in \set{C}_{\hat \imath}
        \iff
        \hat \imath = \argmin_{i \in \{ 1, 2, \dotsc, K \}} \norm{\vect{x}_j - \vect{c}_i}_2^2 .
        \end{equation*}
  \item Recalculate center $\vect{c}_i$ ($i \in \{ 1, 2, \dotsc, K \}$) as the centroid of $\set{C}_i$, that is
        \begin{equation*}
        \vect{c}_i = \dfrac{1}{\card{\set{C}_i}} \sum_{\vect{x}_j \in \set{C}_i} \vect{x}_j .
        \end{equation*}
  \item Repeat steps 2 and 3 until convergence.
\end{enumerate}

K-means is undoubtedly the most widely used partitional clustering algorithm \cite{Jain99,Xu05,Berkhin06,Bock07,Omran07,Ghosh09,Jain10,Xiao12,Hall12}. Its popularity can be attributed to several reasons. First, it is conceptually simple and easy to implement. Virtually every data mining software includes an implementation of it. Second, it is versatile, i.e., almost every aspect of the algorithm (initialization, distance function, termination criterion, etc.) can be modified. This is evidenced by hundreds of publications over the last fifty years that extend k-means in a variety of ways. Third, it has a time complexity that is linear in $N$, $D$, and $K$ (in general, $D \ll N$ and $K \ll N$). For this reason, it can be used to initialize more expensive clustering algorithms such as expectation maximization \cite{Melnykov13}, fuzzy c-means \cite[p.~35]{Bezdek99}, DBSCAN \cite{Dash01}, spectral clustering \cite{vonLuxburg07,Chen11}, ant colony clustering \cite{Monmarche99}, and particle swarm clustering \cite{Merwe03}. Furthermore, numerous sequential \cite{Pelleg99,Kanungo02,Elkan03,Lai09,Hamerly10,Jin11,Drake12} and parallel \cite{Gokhale03,Wu09,Hwang10,Chen10,Bekkerman12,An13,Li13,Kohlhoff13} acceleration techniques are available in the literature. Fourth, it has a storage complexity that is linear in $N$, $D$, and $K$. In addition, there exist disk-based variants that do not require all points to be stored in memory \cite{Bradley98b,Farnstrom00,Ordonez04,Jin06}. Fifth, it is guaranteed to converge \cite{Selim84} at a quadratic rate \cite{Bottou95}. Finally, it is invariant to data ordering, i.e., random shufflings of the data points.

On the other hand, k-means has several significant disadvantages. First, it requires the number of clusters, $K$, to be specified in advance. The value of this parameter can be determined automatically by means of various internal/relative cluster validity measures \cite{Vendramin10,Baarsch12,Arbelaitz13}. Second, it can only detect compact, hyperspherical clusters that are well separated. This can be alleviated by using a more general distance function such as the Mahalanobis distance, which permits the detection of hyperellipsoidal clusters \cite{Mao96,Melnykov14}. Third, due its utilization of the squared Euclidean distance, it is sensitive to noise and outlier points since even a few such points can significantly influence the means of their respective clusters. This can be addressed by outlier pruning \cite{Zhang03} or by using a more robust distance function such as the city-block ($\ell_1$) distance \cite{Spath76,Jajuga87,EstivillCastro04}. Fourth, due to its gradient descent nature, it often converges to a local minimum of the criterion function \cite{Selim84}. For the same reason, it is highly sensitive to the selection of the initial centers \cite{Celebi13}. Adverse effects of improper initialization include empty clusters, slower convergence, and a higher chance of getting stuck in bad local minima \cite{Celebi11}. Fortunately, except for the first two, these drawbacks can be remedied by using an adaptive initialization method (IM).

A large number of IMs have been proposed in the literature \cite{Pena99,He04,Celebi11,deAmorim12,Celebi13}. Unfortunately, many of these have time complexity superlinear in $N$ \cite{Lance67,Astrahan70,Hartigan79,Kaufman90,Likas03,AlDaoud05,Redmond07,AlHasan09,Cao09,Kang09}, which makes them impractical for large data sets (note that k-means itself has linear time complexity). In contrast, linear IMs are often random and/or order-sensitive \cite{Forgy65,Jancey66,MacQueen67,Ball67,Tou74,Spath77,Bradley98a,Arthur07}, which renders their results unreliable. In this study, we investigate the empirical performance of six linear, deterministic (non-random), and order-invariant k-means IMs on a large and diverse collection of data sets from the UCI Machine Learning Repository.

The rest of the chapter is organized as follows. Section \ref{sec_related} presents an overview of linear, deterministic, and order-invariant k-means IMs. Section \ref{sec_exp_setup} describes the experimental setup. Section \ref{sec_exp_results} presents and discusses the experimental results. Finally, Section \ref{sec_conc} gives the conclusions.

\section{Linear, Deterministic, and Order-Invariant K-Means Initialization Methods}
\label{sec_related}

In this study, we focus on IMs that have time complexity linear in $N$. This is because k-means itself has linear complexity, which is perhaps the most important reason for its popularity. Therefore, an IM for k-means should not diminish this advantage of the algorithm. Accordingly, the following six linear, deterministic, and order-invariant IMs are investigated.

The maximin (\texttt{MM}) method \cite{Gonzalez85} chooses the first center $\vect{c}_1$ arbitrarily from the data points and the remaining $(K - 1)$ centers are chosen successively as follows. In iteration $i$ ($i \in \{2, 3, \ldots, K\}$), the $i$th center $\vect{c}_i$ is chosen to be the point with the greatest minimum $\ell_2$ distance to the previously selected $(i - 1)$ centers, i.e., $\vect{c}_1, \vect{c}_2, \ldots, \vect{c}_{i-1}$. This method can be expressed in algorithmic notation as follows:
\begin{enumerate}
  \item Choose the first center $\vect{c}_1$ arbitrarily from the data points.
  \item Choose the next center $\vect{c}_i$ ($i \in \{ 2, 3, \dotsc, K \}$) as the point $\vect{x}_{\hat \jmath}$ that satisfies
        \begin{equation*}
        \hat{\jmath} = \argmax_{j \in \{ 1, 2, \dotsc, N \}} \left( \min_{k \in \{ 1, 2, \dotsc, i - 1 \}} \norm{\vect{x}_j - \vect{c}_k}_2^2 \right) .
        \end{equation*}
  \item Repeat step 2 $(K - 1)$ times.
\end{enumerate}

Despite the fact that it was originally developed as a $2$-approximation to the $K$-center clustering problem\footnote{\label{foot_k_center} Given a set of $N$ points in a metric space, the goal of $K$-center clustering is to find $K$ representative points (centers) such that the maximum distance of a point to a center is minimized \cite[p.~63]{HarPeled11}. A polynomial-time algorithm is said to be a \emph{$\delta$-approximation} algorithm for a minimization problem if for every instance of the problem it delivers a solution whose cost is at most $\delta$ times the cost of the optimal solution ($\delta$ is often referred to as the ``approximation ratio'' or ``approximation factor'')~\cite[p.~xv]{Hochbaum97}.}, \texttt{MM} is commonly used as a k-means initializer\footnote{Interestingly, several authors including Thorndike \cite{Thorndike53}, Casey and Nagy \cite{Casey68}, Batchelor and Wilkins \cite{Batchelor69}, Kennard and Stone \cite{Kennard69}, and Tou and Gonzalez \cite[pp.~92--94]{Tou74} had proposed similar (or even identical) methods decades earlier. Gonzalez \cite{Gonzalez85}, however, was the one to prove the theoretical properties of the method.}. In this study, the first center is chosen to be the centroid of $\set{X}$ given by
\begin{equation}
\bar{\vect{x}} = \dfrac{1}{N} \sum_{j = 1}^N \vect{x}_j .
\end{equation}
Note that $\vect{c}_1 = \bar{\vect{x}}$ gives the optimal SSE when $K = 1$.

Katsavounidis \emph{et al.}'s method (\texttt{KK}) \cite{Katsavounidis94} is identical to \texttt{MM} with the exception that the first center is chosen to be the point with the greatest $\ell_2$ norm\footnote{This choice was motivated by a vector quantization application.}, that is, the point $\vect{x}_{\hat \jmath}$ that satisfies
\begin{equation}
\hat{\jmath} = \argmax_{j \in \{ 1, 2, \dotsc, N \}} \norm{\vect{x}_j}_2^2 .
\end{equation}

The PCA-Part (\texttt{PP}) method \cite{Su07} uses a divisive hierarchical approach based on Principal Component Analysis (PCA)~\cite{Jolliffe02}. Starting from an initial cluster that contains the entire data set $\set{X}$, the method successively selects the cluster with the greatest SSE and divides it into two subclusters using a hyperplane that passes through the cluster centroid and is orthogonal to the principal eigenvector of the cluster covariance matrix. This iterative cluster selection and splitting procedure is repeated $(K - 1)$ times. The final centers are then given by the centroids of the resulting $K$ subclusters. This method can be expressed in algorithmic notation as follows:
\begin{enumerate}
  \item Let $\set{C}_i$ be the cluster with the greatest SSE and $\vect{c}_i$ be the centroid of this cluster. In the first iteration, $\set{C}_1 = \set{X}$ and $\vect{c}_1 = \bar{\vect{x}}$.
  \item Let $p$ be the projection of $\vect{c}_i$ on the principal eigenvector $\vect{v}_i$ of $\set{C}_i$, i.e., $p = \vect{c}_i \cdot \vect{v}_i$, where `$\cdot$' denotes the dot product.
  \item Divide $\set{C}_i$ into two subclusters $\set{C}_{i_1}$ and $\set{C}_{i_2}$ according to the following rule: For any $\vect{x}_j \in \set{C}_i$, if $\vect{x}_j \cdot \vect{v}_i \leq p$, then assign $\vect{x}_j$ to $\set{C}_{i_1}$; otherwise, assign it to $\set{C}_{i_2}$.
  \item Repeat steps 1--3 $(K -1)$ times.
\end{enumerate}

The Var-Part (\texttt{VP}) method \cite{Su07} is an approximation to \texttt{PP}, where, in each iteration, the covariance matrix of the cluster to be split is assumed to be diagonal. In this case, the splitting hyperplane is orthogonal to the coordinate axis with the greatest variance. In other words, the only difference between \texttt{VP} and \texttt{PP} is the choice of the projection axis.

\begin{figure}[!ht]
\centering
 \subfigure[Input data set]{\includegraphics[width=0.48\columnwidth]{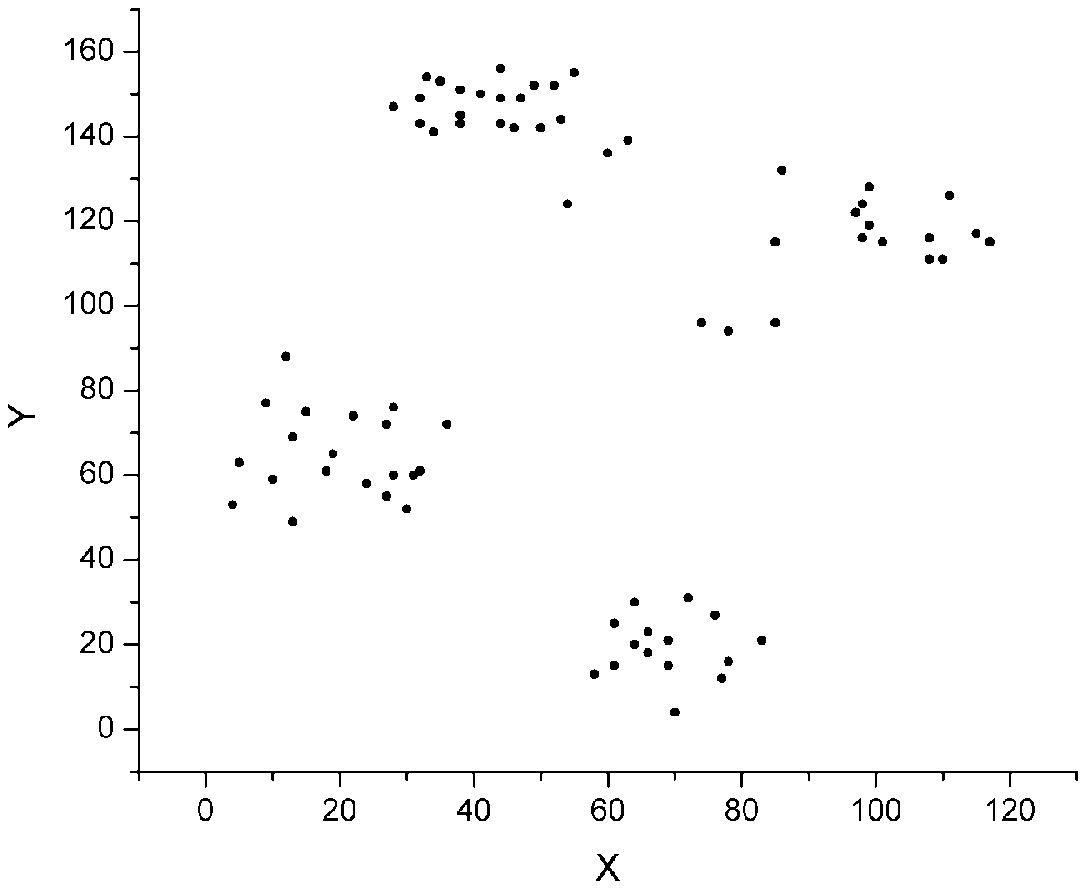}}
 \subfigure[Iteration 1]{\includegraphics[width=0.48\columnwidth]{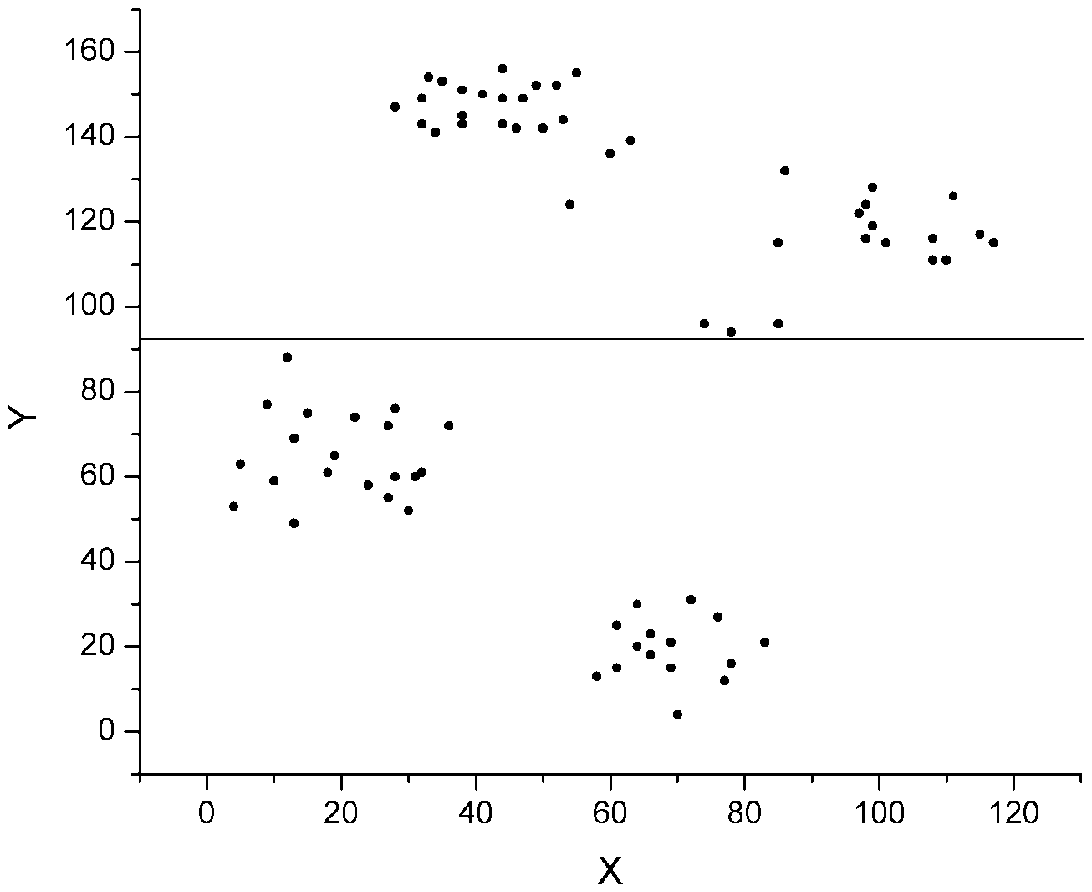}}
 \\
 \subfigure[Iteration 2]{\includegraphics[width=0.48\columnwidth]{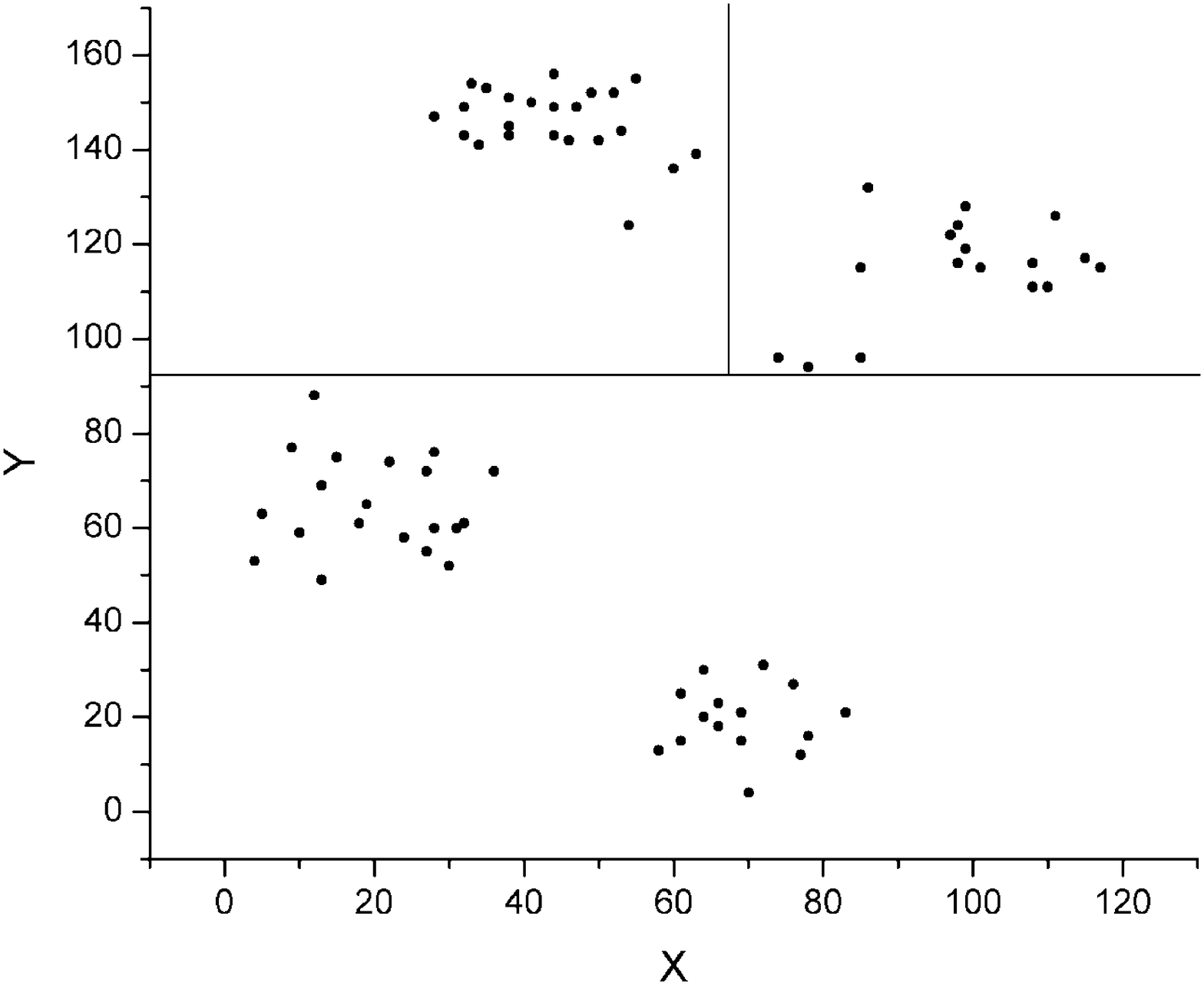}}
 \subfigure[Iteration 3]{\label{part_d} \includegraphics[width=0.48\columnwidth]{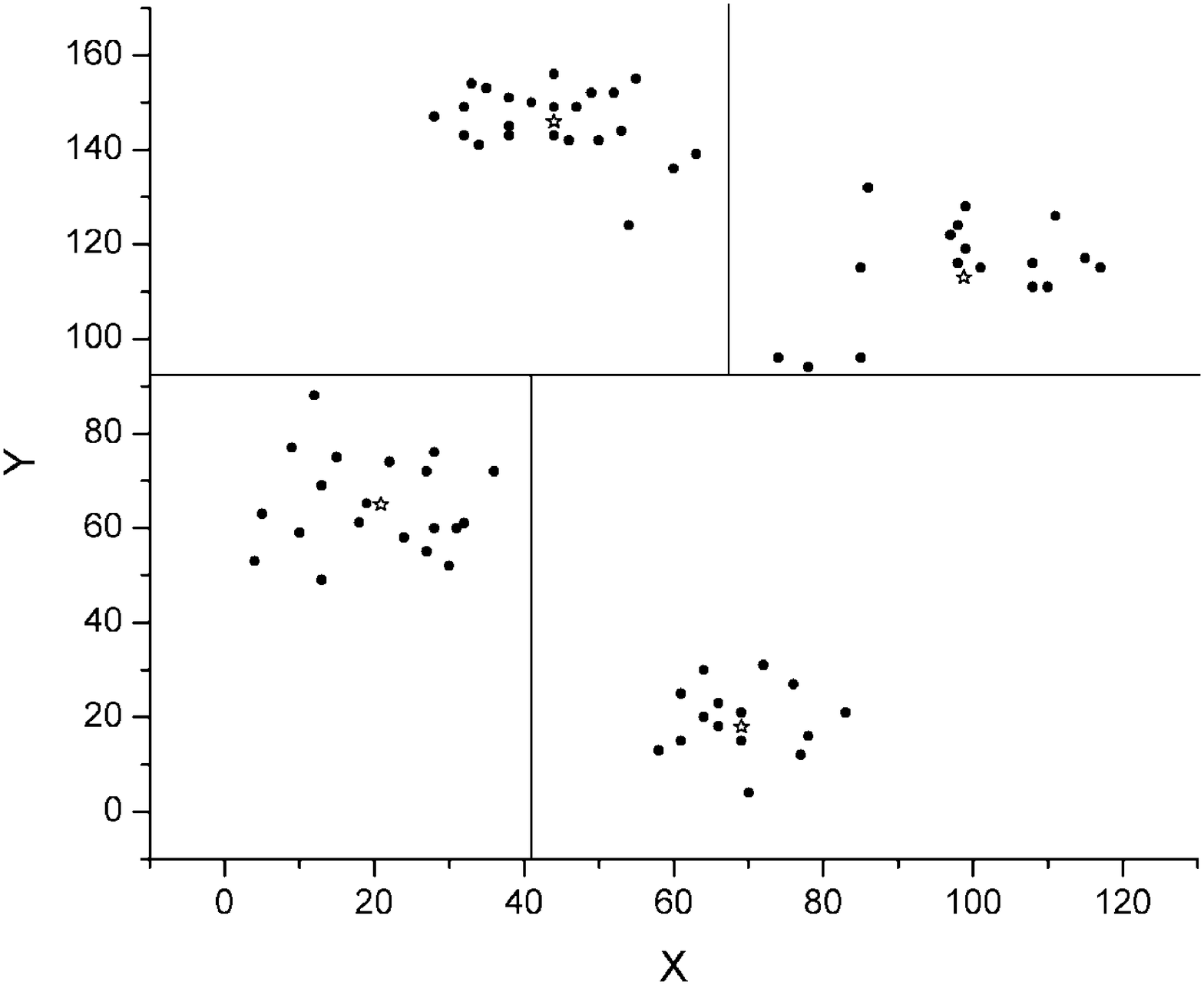}}
 \caption{Illustration of Var-Part on the Ruspini data set}
 \label{fig_varpart_ruspini}
\end{figure}

Figure \ref{fig_varpart_ruspini} \cite{Celebi12} illustrates \texttt{VP} on a toy data set with four natural clusters \cite{Ruspini70}\cite[p.~100]{Kaufman90}. In iteration $1$, the initial cluster that contains the entire data set is split into two subclusters along the Y axis using a line (i.e., a one-dimensional hyperplane) passing through the mean point ($92.026667$). Between the resulting two clusters, the one above the line has a greater SSE. In iteration $2$, this cluster is thus split along the X axis at the mean point ($66.975000$). In the final iteration, the cluster with the greatest SSE, i.e., the bottom cluster, is split along the X axis at the mean point ($41.057143$). In Figure \ref{part_d}, the centroids of the final four clusters are denoted by stars.

The maxisum (\texttt{MS}) method\footnote{Eri\c{s}o\u{g}lu \emph{et al.}, Pattern Recognition Letters \textbf{32}(14), 1701--1705 (2011)} is a recent modification of \texttt{MM}. It can be expressed in algorithmic notation as follows:
\begin{enumerate}
  \item Determine the attribute with the greatest absolute \emph{coefficient of variation} (ratio of the standard deviation to the mean), that is, the attribute $\vect{x}_{.d_1}$ that satisfies
  \begin{equation*}
  d_1 = \argmax_{d \in \{ 1, 2, \dotsc, D \}} \abs{\dfrac{s_d}{m_d}} ,
  \end{equation*}
  where
  \begin{equation*}
  m_d = \dfrac{1}{N} \sum_{j = 1}^N x_{jd}
  \end{equation*}
  and
  \begin{equation*}
  s_d^2 = \dfrac{1}{N - 1} \sum_{j = 1}^N (x_{jd} - m_d)^2
  \end{equation*}
  denote the mean and variance of the $d$th attribute $\vect{x}_{.d}$, respectively.
  \item Determine the attribute with the least \emph{Pearson product-moment correlation} with $\vect{x}_{.d_1}$, that is, the attribute $\vect{x}_{.d_2}$ that satisfies
  \begin{equation}
  \label{eq_corr}
  d_2 =
  \argmin_{d \in \{ 1, 2, \dotsc, D \}}
  \dfrac
  {\sum\limits_{j = 1}^N {(x_{jd_1} - m_{d_1})(x_{jd} - m_d)}}
  {
  \sqrt{\sum\limits_{j = 1}^N {(x_{jd_1} - m_{d_1})^2}}
  \sqrt{\sum\limits_{j = 1}^N {(x_{jd} - m_d)^2}}
  }
  .
  \end{equation}

  Note that since we calculated the mean and standard deviation of each attribute in step 1, the following expression can be used in place of \eqref{eq_corr} to save computational time:
  \begin{equation}
  d_2 =
  \argmin_{d \in \{ 1, 2, \dotsc, D \}}
  \sum\limits_{j = 1}^N
  {
  \left( \dfrac{x_{jd_1} - m_{d_1}}{s_{d_1}} \right)
  \left( \dfrac{x_{jd} - m_d}{s_d} \right)
  }
  .
  \end{equation}
  \item Let $\set{Y} = \{ \vect{y}_1, \vect{y}_2, \dotsc, \vect{y}_N \} \subset \mathbb{R}^2$ be the projection of $\set{X} = \{ \vect{x}_1, \vect{x}_2, \dotsc, \vect{x}_N \} \subset \mathbb{R}^D$ onto the two-dimensional subspace determined in steps 1 and 2. In other words, $\vect{y}_j = \left( x_{jd_1},x_{jd_2} \right)$ for $j \in \{ 1, 2, \dotsc, N \}$.
  \item Choose the first center $\vect{c}_1$ as the point farthest from the centroid $\bar{\vect{y}}$ of $\set{Y}$ with respect to $\ell_2$ distance, that is, the point $\vect{y}_{\hat \jmath}$ that satisfies
  \begin{equation*}
  \hat{\jmath} = \argmax_{j \in \{ 1, 2, \dotsc, N \}} \norm{\vect{y}_j - \bar{\vect{y}}}_2^2 .
  \end{equation*}
  \item Choose the next center $\vect{c}_i$ ($i \in \{ 2, 3, \ldots, K \}$) as the point with the greatest cumulative $\ell_2$ distance from the previously selected $(i - 1)$ centers, that is, the point $\vect{y}_{\hat \jmath}$ that satisfies
  \begin{equation*}
  \hat{\jmath} = \argmax_{j \in \{ 1, 2, \dotsc, N \}} \sum_{k = 1}^{i - 1} \norm{\vect{y}_j - \vect{c}_k}_2 .
  \end{equation*}
  \item Repeat step 5 $(K - 1)$ times.
\end{enumerate}

Note that steps 1 and 2 above provide rough approximations to the first two PCs and that steps 4 and 5 are performed in the two-dimensional subspace spanned by the attributes determined in steps 1 and 2.

Clearly, \texttt{MS} is a derivative of \texttt{MM}. Differences between the two methods are as follows:
\begin{itemize}
\renewcommand{\labelitemi}{$\triangleright$}
  \item \texttt{MM} chooses the first center arbitrarily from the data points, whereas \texttt{MS} chooses it to be the point farthest from the mean of the projected data set.
  \item \texttt{MM} chooses the remaining $(K - 1)$ centers iteratively based on their minimum distance from the previously selected centers, whereas \texttt{MS} uses a cumulative distance criterion. Note that while the selection criterion used in \texttt{MM} provides an approximation guarantee of factor $2$ for the $K$-center clustering problem (see footnote \ref{foot_k_center} on page \pageref{foot_k_center}), it is unclear whether or not \texttt{MS} offers any approximation guarantees.
  \item \texttt{MM} performs the distance calculations in the original $D$-dimensional space, whereas \texttt{MS} works in a two-dimensional subspace. A serious drawback of the projection operation employed in \texttt{MS} is that the method disregards all attributes but two and therefore is likely to be effective only for data sets in which the variability is mostly on two dimensions. Unfortunately, the motivation behind this particular projection scheme is not given by Eri\c{s}o\u{g}lu \emph{et al.}
\end{itemize}

Interestingly, \texttt{MS} also bears a striking resemblance to a method proposed by DeSarbo \emph{et al.} \cite{DeSarbo84} almost three decades earlier. The latter method differs from the former in two ways. First, it works in the original $D$-dimensional space. Second, it chooses the first two centers as the pair of points with the greatest $\ell_2$ distance. Unfortunately, the determination of the first two centers in this method leads to a time complexity quadratic in $N$. Therefore, this method was not included in the experiments. More recently, Glasbey \emph{et al.} \cite{Glasbey06} mentioned a very similar method  within the context of color palette design.

We also experimented with a modified version of the \texttt{MS} method (\texttt{MS+}), which is identical to \texttt{MS} with the exception that there is no projection involved. In other words, \texttt{MS+} operates in the original $D$-dimensional space.

For a comprehensive overview of these methods and others, the reader is referred to a recent article by Celebi \emph{et al.} \cite{Celebi13}. It should be noted that, in this study, we do not attempt to compare a mix of deterministic and random IMs. Instead, we focus on deterministic methods for two main reasons. First, these methods are generally computationally more efficient as they need to be executed only once. In contrast, random methods are inherently unreliable in that the quality of their results is unpredictable and thus it is common practice to perform multiple runs of such methods and take the output of the run\footnote{Each `run' of a random IM involves the execution of the IM itself followed by that of the clustering algorithm, e.g., k-means.} that produces the best objective function value. Second, several studies \cite{Su07,Celebi12,Celebi13} demonstrated that despite the fact that they are executed only once, some deterministic methods are highly competitive with well-known and effective random methods such as Bradley and Fayyad's method \cite{Bradley98a} and k-means++ \cite{Arthur07}.

\section{Experimental Setup}
\label{sec_exp_setup}
\subsection{Data Set Descriptions}
The experiments were performed on $24$ commonly used data sets from the UCI Machine Learning Repository \cite{Frank14}. Table \ref{tab_data_set} gives the data set descriptions. For each data set, the number of clusters ($K$) was set equal to the number of classes ($K'$), as commonly seen in the related literature \cite{He04,Arthur07,Su07,Redmond07,AlHasan09,Cao09,Kang09,Onoda12,Celebi12,Celebi13,Gingles14}.

\begin{table*}[ht]
\renewcommand{\arraystretch}{1.2}
\centering
\caption{\label{tab_data_set} Data Set Descriptions ($N$: \# points, $D$: \# attributes, $K'$: \# classes)}
\begin{tabular}{@{\extracolsep{\fill}} llrrr}
\hline
ID & Data Set & $N$ & $D$ & $K'$\\
\hline
\hline
01 & Breast Cancer Wisconsin (Original) & 683 & 9 & 2\\
02 & Breast Tissue & 106 & 9 & 6\\
03 & Ecoli & 336 & 7 & 8\\
04 & Steel Plates Faults & 1,941 & 27 & 7\\
05 & Glass Identification & 214 & 9 & 6\\
06 & Heart Disease (Cleveland) & 297 & 13 & 5\\
07 & Ionosphere & 351 & 34 & 2\\
08 & Iris (Bezdek) & 150 & 4 & 3\\
09 & ISOLET & 7,797 & 617 & 26\\
10 & Landsat Satellite (Statlog) & 6,435 & 36 & 6\\
11 & Letter Recognition & 20,000 & 16 & 26\\
12 & Multiple Features (Fourier) & 2,000 & 76 & 10\\
13 & Libras Movement & 360 & 90 & 15\\
14 & Optical Digits & 5,620 & 64 & 10\\
15 & Page Blocks Classification & 5,473 & 10 & 5\\
16 & Pen Digits & 10,992 & 16 & 10\\
17 & Person Activity & 164,860 & 3 & 11\\
18 & Image Segmentation & 2,310 & 19 & 7\\
19 & Shuttle (Statlog) & 58,000 & 9 & 7\\
20 & Spambase & 4,601 & 57 & 2\\
21 & Vertebral Column & 310 & 6 & 3\\
22 & Wall-Following Robot Navigation & 5,456 & 24 & 4\\
23 & Wine & 178 & 13 & 3\\
24 & Yeast & 1,484 & 8 & 10\\
\hline
\end{tabular}
\end{table*}

\subsection{Attribute Normalization}
Normalization is a common preprocessing step in clustering that is necessary to prevent attributes with large variability from dominating the distance calculations and also to avoid numerical instabilities in the computations. Two commonly used normalization schemes are linear scaling to unit range (min-max normalization) and linear scaling to unit variance (z-score normalization). Several studies revealed that the former scheme is preferable to the latter since the latter is likely to eliminate valuable between-cluster variation \cite{Milligan88,Gnanadesikan95,Gnanadesikan07}. As a result, we used the min-max normalization scheme to map the attributes of each data set to the $[0,1]$ interval.

\subsection{Performance Criteria}
The performance of the IMs was quantified using two effectiveness (quality) and one efficiency (speed) criteria:
\begin{itemize}
\renewcommand{\labelitemi}{$\triangleright$}
 \item \textbf{Initial SSE (IS)}: This is the SSE value calculated after the initialization phase, before the clustering phase. It gives us a measure of the effectiveness of an IM by itself.
 \item \textbf{Final SSE (FS)}: This is the SSE value calculated after the clustering phase. It gives us a measure of the effectiveness of an IM when its output is refined by k-means. Note that this is the objective function of the k-means algorithm, i.e., \eqref{eq_sse}.
 \item \textbf{Number of Iterations (NI)}: This is the number of iterations that k-means requires until reaching convergence when initialized by a particular IM. It is an efficiency measure independent of programming language, implementation style, compiler, and CPU architecture. Note that we do not report CPU time measurements since on most data sets that we tested each of the six IMs completed within a few milliseconds (gcc v4.4.5, Intel Core i7-3960X 3.30GHz).
\end{itemize}

The convergence of k-means was controlled by the disjunction of two criteria: the number of iterations reaches a maximum of $100$ or the relative improvement in SSE between two consecutive iterations drops below a threshold \cite{Linde80}, i.e., $\left( {\text{SSE}_{i - 1} - \text{SSE}_i} \right) / \text{SSE}_i \leq \epsilon$, where $\text{SSE}_i$ denotes the SSE value at the end of the $i$th ($i \in \{ 2, \dotsc, 100 \}$) iteration. The convergence threshold was set to $\epsilon = 10^{-6}$.

\section{Experimental Results and Discussion}
\label{sec_exp_results}
Tables \ref{tab_init_sse}--\ref{tab_ni} give the performance measurements for each method (the best values are \ul{underlined}). Since the number of iterations fall within $[0,100]$, we can directly obtain descriptive statistics such as minimum, maximum, mean, and median for this criterion over the $24$ data sets. In contrast, initial/final SSE values are unnormalized and therefore incomparable across different data sets. In order to circumvent this problem, for each data set, we calculated the percent SSE of each method relative to the worst (greatest) SSE. For example, it can be seen from Table \ref{tab_init_sse} that on the Breast Cancer Wisconsin data set the initial SSE of \texttt{MM} is $498$, whereas the worst initial SSE on the same data set is $596$ and thus the ratio of the former to the latter is $0.836$. This simply means that on this data set \texttt{MM} obtains $100(1-0.836)\approx16$\% better initial SSE than the worst method, \texttt{KK}. Table \ref{tab_overall} gives the summary statistics for the normalized initial/final SSE's obtained in this manner and those for the number of iterations. As usual, \emph{min} (minimum) and \emph{max} (maximum) represent the \emph{best} and \emph{worst} case performance, respectively. \emph{Mean} represents the \emph{average} case performance, whereas median quantifies the \emph{typical} performance of a method without regard to outliers. For example, with respect to the initial SSE criterion, \texttt{PP} performs, on the \emph{average}, about $100-21.46 \approx 79$\% better than the worst method.

\begin{table*}[ht]
\renewcommand{\arraystretch}{1.2}
\centering
\caption{\label{tab_init_sse} Initial SSE comparison of the initialization methods}
\begin{tabular}{@{\extracolsep{\fill}} lrrrrrr}
\hline
ID & \texttt{MM} & \texttt{KK} & \texttt{VP} & \texttt{PP} & \texttt{MS} & \texttt{MS+}\\
\hline
\hline
01 & 498 & 596 & 247 & \ul{240} & 478 & 596\\
02 & 19 & 18 & \ul{8} & \ul{8} & 50 & 21\\
03 & 48 & 76 & 20 & \ul{19} & 104 & 68\\
04 & 2817 & 3788 & \ul{1203} & 1262 & 5260 & 4627\\
05 & 45 & 117 & 21 & \ul{20} & 83 & 132\\
06 & 409 & 557 & \ul{249} & 250 & 773 & 559\\
07 & 827 & 1791 & 632 & \ul{629} & 3244 & 3390\\
08 & 18 & 23 & \ul{8} & \ul{8} & 42 & 42\\
09 & 221163 & 298478 & 145444 & \ul{124958} & 368510 & 318162\\
10 & 4816 & 7780 & \ul{2050} & 2116 & 7685 & 11079\\
11 & 5632 & 7583 & 3456 & \ul{3101} & 12810 & 14336\\
12 & 4485 & 7821 & 3354 & \ul{3266} & 7129 & 8369\\
13 & 1023 & 1114 & 628 & \ul{592} & 1906 & 1454\\
14 & 25291 & 36691 & 17476 & \ul{15714} & 43169 & 42213\\
15 & 635 & 2343 & 300 & \ul{230} & 1328 & 7868\\
16 & 12315 & 16159 & 5947 & \ul{5920} & 17914 & 16104\\
17 & 5940 & 7196 & \ul{1269} & 1468 & 42475 & 50878\\
18 & 1085 & 1617 & 472 & \ul{416} & 3071 & 1830\\
19 & 1818 & 14824 & 316 & \ul{309} & 26778 & 28223\\
20 & \ul{772} & 13155 & 782 & 783 & 5101 & 13155\\
21 & 37 & 103 & 23 & \ul{20} & 83 & 103\\
22 & 11004 & 21141 & 8517 & \ul{7805} & 19986 & 20122\\
23 & 87 & 185 & \ul{51} & 53 & 153 & 212\\
24 & 115 & 261 & 77 & \ul{63} & 209 & 658\\
\hline
\end{tabular}
\end{table*}

\begin{table*}[ht]
\renewcommand{\arraystretch}{1.2}
\centering
\caption{\label{tab_final_sse} Final SSE comparison of the initialization methods}
\begin{tabular}{@{\extracolsep{\fill}} lrrrrrr}
\hline
ID & \texttt{MM} & \texttt{KK} & \texttt{VP} & \texttt{PP} & \texttt{MS} & \texttt{MS+}\\
\hline
\hline
01 & \ul{239} & \ul{239} & \ul{239} & \ul{239} & \ul{239} & \ul{239}\\
02 & \ul{7} & \ul{7} & \ul{7} & \ul{7} & 11 & 10\\
03 & 19 & 20 & \ul{17} & 18 & 40 & 20\\
04 & 1331 & 1329 & \ul{1167} & 1168 & 1801 & 1376\\
05 & 23 & 23 & \ul{19} & \ul{19} & 31 & 22\\
06 & 249 & 249 & 248 & \ul{243} & 276 & 253\\
07 & 826 & \ul{629} & \ul{629} & \ul{629} & \ul{629} & \ul{629}\\
08 & \ul{7} & \ul{7} & \ul{7} & \ul{7} & \ul{7} & \ul{7}\\
09 & 135818 & 123607 & 118495 & \ul{118386} & 174326 & 121912\\
10 & \ul{1742} & \ul{1742} & \ul{1742} & \ul{1742} & \ul{1742} & \ul{1742}\\
11 & 2749 & 2783 & \ul{2735} & 2745 & 4520 & 3262\\
12 & 3316 & 3284 & \ul{3137} & 3214 & 3518 & 3257\\
13 & 502 & 502 & 502 & \ul{486} & 783 & 530\\
14 & 14679 & 14649 & \ul{14581} & 14807 & 21855 & \ul{14581}\\
15 & 230 & 295 & 227 & \ul{215} & 230 & 310\\
16 & 5049 & \ul{4930} & \ul{4930} & 5004 & 7530 & 5017\\
17 & 1195 & 1195 & 1182 & \ul{1177} & 1226 & 1192\\
18 & 433 & 443 & 410 & \ul{405} & 745 & 446\\
19 & 726 & 658 & \ul{235} & 274 & 728 & 496\\
20 & \ul{765} & \ul{765} & 778 & 778 & 778 & \ul{765}\\
21 & 23 & 23 & \ul{19} & \ul{19} & 23 & 23\\
22 & \ul{7772} & \ul{7772} & 7774 & 7774 & \ul{7772} & \ul{7772}\\
23 & 63 & \ul{49} & \ul{49} & \ul{49} & \ul{49} & \ul{49}\\
24 & 61 & 61 & 69 & \ul{59} & 60 & 63\\
\hline
\end{tabular}
\end{table*}

\begin{table*}[ht]
\renewcommand{\arraystretch}{1.2}
\centering
\caption{\label{tab_ni} Number of iterations comparison of the initialization methods}
\begin{tabular}{@{\extracolsep{\fill}} lrrrrrr}
\hline
ID & \texttt{MM} & \texttt{KK} & \texttt{VP} & \texttt{PP} & \texttt{MS} & \texttt{MS+}\\
\hline
\hline
01 & 8 & 7 & \ul{4} & \ul{4} & 7 & 7\\
02 & 7 & 6 & 6 & 7 & 9 & \ul{4}\\
03 & 14 & 12 & 17 & 7 & \ul{4} & 10\\
04 & 25 & 16 & \ul{11} & 42 & 12 & 12\\
05 & 6 & \ul{5} & 6 & \ul{5} & 7 & 6\\
06 & 12 & 10 & \ul{3} & 4 & 11 & 16\\
07 & \ul{3} & 6 & \ul{3} & \ul{3} & 7 & 6\\
08 & 6 & 5 & \ul{4} & \ul{4} & 12 & 19\\
09 & \ul{32} & 36 & 82 & 45 & 34 & 81\\
10 & 53 & \ul{17} & 28 & 27 & 24 & 33\\
11 & 72 & \ul{63} & 100 & 83 & 91 & 65\\
12 & 37 & 32 & \ul{14} & 25 & 31 & 29\\
13 & 13 & \ul{7} & 17 & 11 & 18 & 16\\
14 & 36 & 24 & \ul{16} & 22 & 29 & 17\\
15 & 27 & 18 & 25 & 15 & 30 & \ul{12}\\
16 & 19 & 17 & \ul{13} & 17 & 22 & 29\\
17 & \ul{31} & \ul{31} & 100 & 63 & 91 & 53\\
18 & 31 & \ul{9} & 10 & 18 & 16 & 22\\
19 & 22 & \ul{8} & 30 & 16 & 14 & 9\\
20 & \ul{5} & \ul{5} & 9 & 10 & 11 & \ul{5}\\
21 & 11 & 10 & 10 & 9 & \ul{8} & 10\\
22 & 24 & 14 & 20 & \ul{8} & 20 & 19\\
23 & 9 & 7 & \ul{5} & 7 & 7 & 8\\
24 & 73 & 43 & 33 & \ul{21} & 71 & 49\\
\hline
\end{tabular}
\end{table*}

\begin{table*}[ht]
\renewcommand{\arraystretch}{1.2}
\centering
\caption{\label{tab_overall} Summary statistics for Tables \ref{tab_init_sse}--\ref{tab_ni}}
\begin{tabular}{@{\extracolsep{\fill}} clrrrrrr}
\hline
Criterion & Statistic & \texttt{MM} & \texttt{KK} & \texttt{VP} & \texttt{PP} & \texttt{MS} & \texttt{MS+}\\
\hline
\hline
\multirow{6}{*}{IS} & Min & 5.87 & 14.14 & 1.12 & 1.10 & 16.88 & 42.00\\
 & Q1 & 29.24 & 52.74 & 15.64 & 14.35 & 76.19 & 87.15\\
 & Median & 41.95 & 72.04 & 20.78 & 19.26 & 94.71 & 100.00\\
 & Q3 & 53.57 & 89.42 & 33.07 & 32.69 & 100.00 & 100.00\\
 & Max & 83.56 & 100.00 & 41.44 & 40.27 & 100.00 & 100.00\\
 & Mean & 40.28 & 69.03 & 22.55 & 21.46 & 83.16 & 90.53\\
\hline
\multirow{6}{*}{FS} & Min & 47.50 & 50.00 & 32.28 & 37.64 & 74.19 & 50.00\\
 & Q1 & 67.11 & 66.25 & 63.87 & 62.85 & 100.00 & 69.03\\
 & Median & 89.31 & 83.09 & 74.69 & 72.75 & 100.00 & 84.34\\
 & Q3 & 99.99 & 97.90 & 98.21 & 93.68 & 100.00 & 99.15\\
 & Max & 100.00 & 100.00 & 100.00 & 100.00 & 100.00 & 100.00\\
 & Mean & 83.21 & 81.56 & 76.23 & 75.77 & 96.46 & 82.68\\
\hline
\multirow{6}{*}{NI} & Min & 3.00 & 5.00 & 3.00 & 3.00 & 4.00 & 4.00\\
 & Q1 & 8.50 & 7.00 & 6.00 & 7.00 & 8.50 & 8.50\\
 & Median & 20.50 & 11.00 & 13.50 & 13.00 & 15.00 & 16.00\\
 & Q3 & 31.50 & 21.00 & 26.50 & 23.50 & 29.50 & 29.00\\
 & Max & 73.00 & 63.00 & 100.00 & 83.00 & 91.00 & 81.00\\
 & Mean & 24.00 & 17.00 & 23.58 & 19.71 & 24.42 & 22.38\\
\hline
\end{tabular}
\end{table*}

\begin{figure*}[!ht]
\centering
 \subfigure[Normalized Initial SSE]{\label{fig_boxplot_is} \includegraphics[width=0.48\columnwidth]{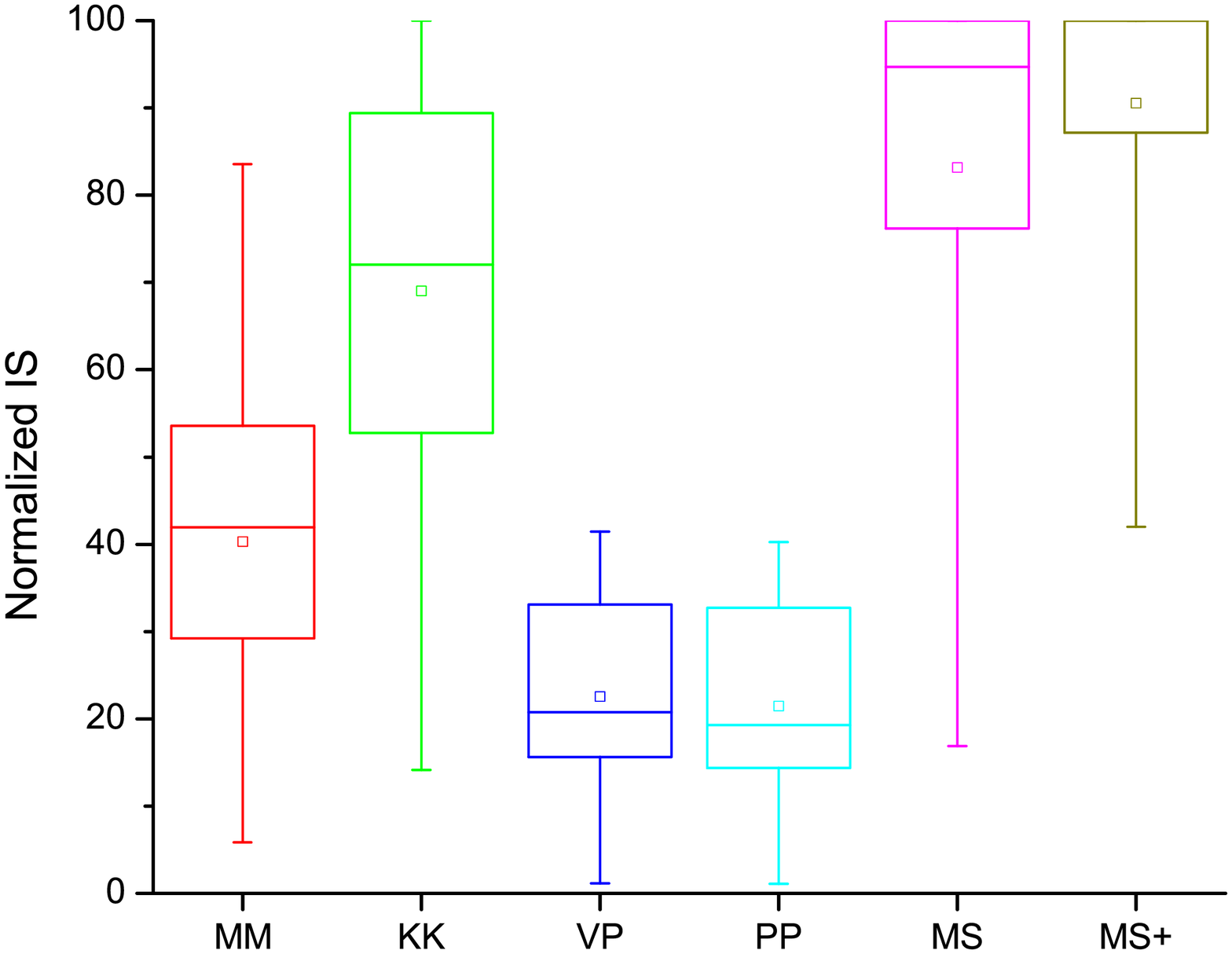}}
 \subfigure[Normalized Final SSE]{\label{fig_boxplot_fs} \includegraphics[width=0.48\columnwidth]{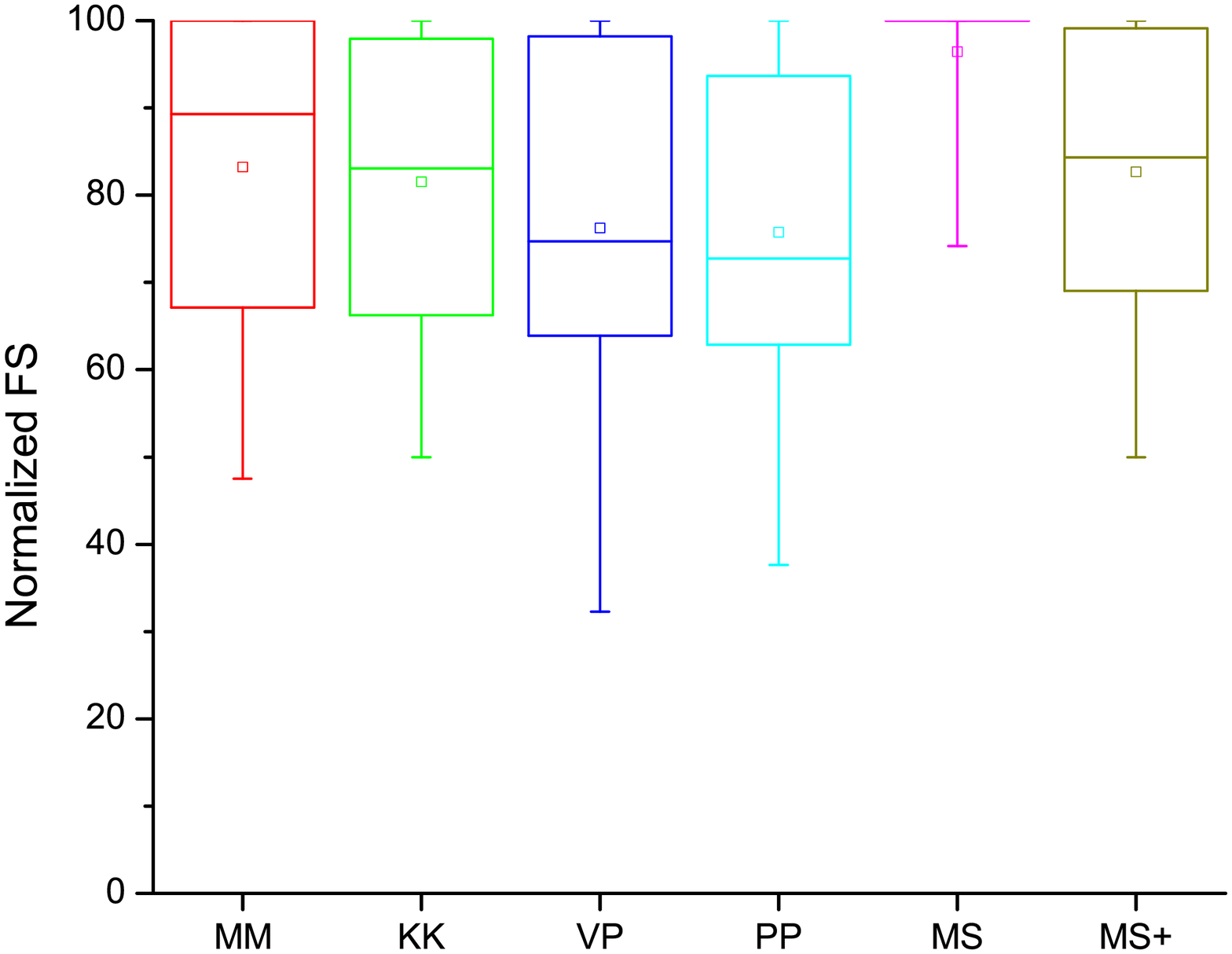}}
 \subfigure[Number of Iterations]{\includegraphics[width=0.48\columnwidth]{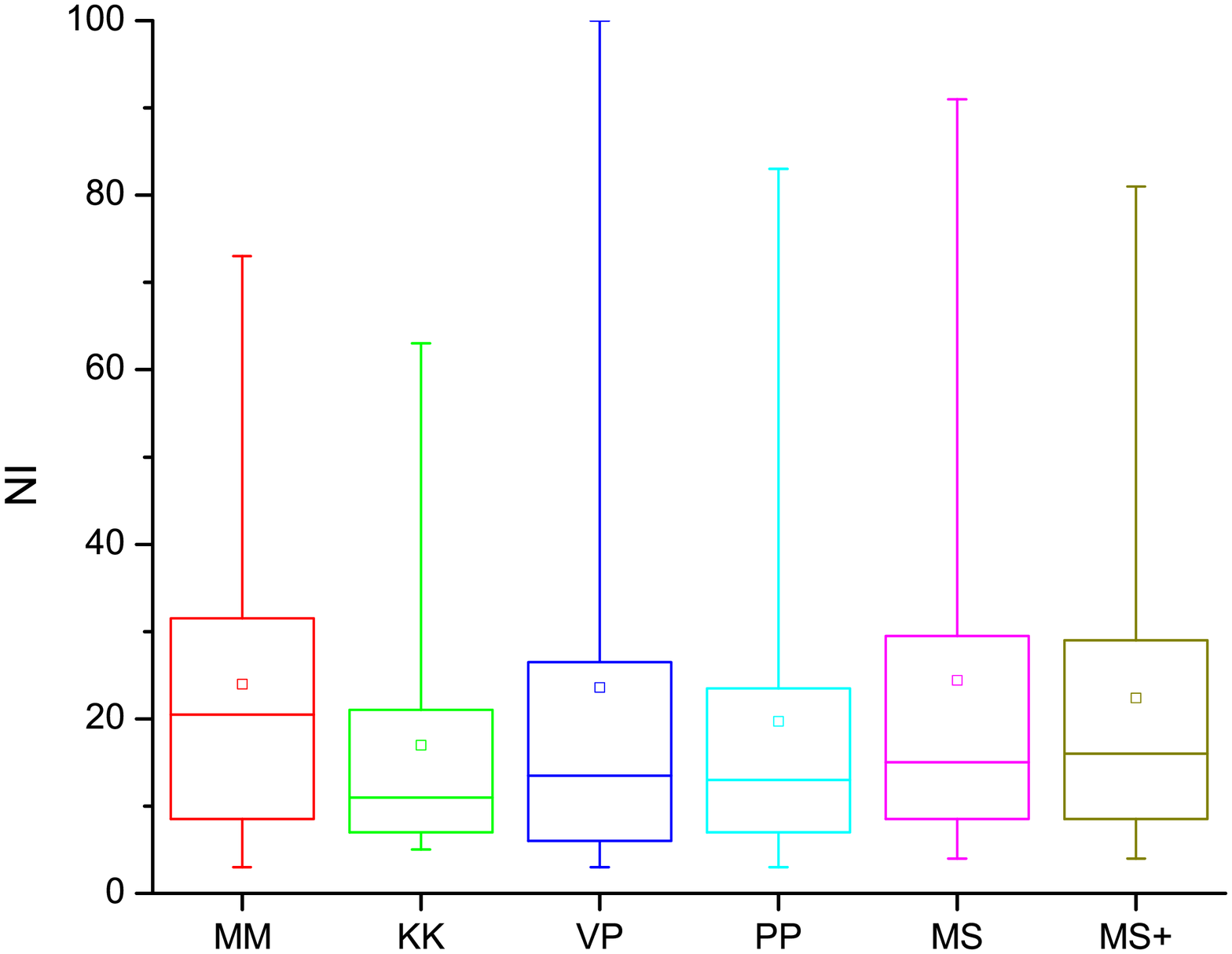}}
 \caption{Box plots for the performance criteria}
 \label{fig_boxplots}
\end{figure*}

For convenient visualization, Figure \ref{fig_boxplots} shows the box plots that depict the five-number summaries (minimum, $25$th percentile, median, $75$th percentile, and maximum) for the normalized initial/final SSE's calculated in the aforementioned manner and the five-number summary for the number of iterations. Here, the bottom and top end of the whiskers of a box represent the \emph{minimum} and \emph{maximum}, respectively, whereas the bottom and top of the box itself are the $25$th percentile (\emph{Q1}) and $75$th percentile (\emph{Q3}), respectively. The line that passes through the box is the $50$th percentile (\emph{Q2}), i.e., the \emph{median}, while the small square inside the box denotes the \emph{mean}. 

With respect to effectiveness, the following observations can be made:
\begin{itemize}
\renewcommand{\labelitemi}{$\triangleright$}
  \item \texttt{VP} and \texttt{PP} performed very similarly with respect to both initial and final SSE.
  \item On $23$ (out of $24$) data sets, \texttt{VP} and \texttt{PP} obtained the two best initial SSE's. Therefore, in applications where an approximate clustering of the data set is desired, these hierarchical methods should be used.
  \item On $23$ data sets, either \texttt{MS} or \texttt{MS+} obtained the worst initial SSE. In fact, on one data set (\#$19$, Shuttle), these methods gave respectively $86.7$ and $91.3$ times worse initial SSE than the best method, \texttt{PP}.
  \item On $20$ data sets, \texttt{VP} and \texttt{PP} obtained the two best final SSE's. Since final SSE is the objective function of k-means, from an optimization point of view, these two methods are the best IMs.
  \item On $16$ data sets, \texttt{MS} obtained the worst final SSE. In fact, on one data set (\#$19$, Shuttle), \texttt{MS} gave $3.1$ times worse final SSE than the best method, \texttt{VP}.
  \item A comparison between Figs.\ \ref{fig_boxplot_is} and \ref{fig_boxplot_fs} reveals that there is significantly less variation among the IMs with respect to final SSE compared to initial SSE. In other words, the performance of the IMs is more homogeneous with respect to final SSE. This was expected because, being a local optimization procedure, k-means can take two disparate initial configurations to similar (or, in some cases, even identical) local minima. Nevertheless, as Tables \ref{tab_init_sse} and \ref{tab_final_sse} show, \texttt{VP} and \texttt{PP} consistently performed well, whereas \texttt{MS}/\texttt{MS+} generally performed poorly.
\end{itemize}

With respect to computational efficiency, the following observations can be made:
\begin{itemize}
\renewcommand{\labelitemi}{$\triangleright$}
  \item An \emph{average} (or \emph{typical}) run of \texttt{KK} lead to the fastest k-means convergence.
  \item An \emph{average} (or \emph{typical}) run of \texttt{PP} lead to the second fastest k-means convergence.
  \item An \emph{average} run of \texttt{MS} lead to the slowest k-means convergence.
  \item A \emph{typical} run of \texttt{MM} lead to the slowest k-means convergence.
\end{itemize}

In summary, our experiments showed that \texttt{VP} and \texttt{PP} performed very similarly with respective to both effectiveness criteria and they outperformed the remaining four methods by a large margin. The former method has a time complexity of $\mathcal{O}(ND)$, whereas the latter one has a complexity of  $\mathcal{O}(ND^2)$ when implemented using the power method \cite{Hotelling36}. Therefore, on high dimensional data sets, the former method might be preferable. On the other hand, on low dimensional data sets, the latter method might be preferable as it often leads to faster k-means convergence. The main disadvantage of these two methods is that they are more complicated to implement due to their hierarchical formulation. As for the remaining four methods, when compared to \texttt{MM}, \texttt{KK} was significantly worse in terms of initial SSE, slightly better in terms of final SSE, and significantly better in terms of number of iterations. Interestingly, despite its similarities with \texttt{MM}, the most recent method that we examined, i.e., \texttt{MS}, often gave the worst results. It was also demonstrated that by eliminating the $2$-dimensional projection step, the performance of \texttt{MS} can be substantially improved with respect to final SSE. This, however, comes at the expense of a performance degradation with respect to initial SSE. Consequently, in either of its forms, the \texttt{MS} method rediscovered recently by Eri\c{s}o\u{g}lu \emph{et al.} does not appear to outperform the classical \texttt{MM} method or the more recent hierarchical methods \texttt{VP} and \texttt{PP}. This is not surprising given that \texttt{MS} can easily choose two nearby points as centers provided that they each have a large cumulative distance to the remaining centers \cite{Glasbey06}.

\section{Conclusions}
\label{sec_conc}
In this chapter we examined six linear, deterministic, and order-invariant methods used for the initialization of the k-means clustering algorithm. These included the popular maximin method and three of its variants and two relatively unknown divisive hierarchical methods. Experiments on a large and diverse collection of real-world data sets from the UCI Machine Learning Repository demonstrated that the hierarchical methods outperform the remaining four methods with respect to two objective effectiveness criteria. These hierarchical methods can be used to initialize k-means effectively, particularly in time-critical applications that involve large data sets. Alternatively, they can be used as approximate clustering algorithms without additional k-means refinement. Our experiments also revealed that the most recent variant of the maximin method performs surprisingly poorly.

\section{Acknowledgments}
This work was supported by a grant from the US National Science Foundation (1117457).

\bibliographystyle{spmpsci}
\bibliography{KMeans_Initialization_Bib}

\end{document}